\documentclass{article}

\usepackage[margin=1in]{geometry}
\usepackage{booktabs}
\usepackage{graphicx}
\usepackage{subcaption}
\usepackage{pgfplots}
\pgfplotsset{compat=1.18}
\graphicspath{{figs/}} % if your PNGs are in figs/
\usepackage[title]{appendix} % For formatting the appendix
\usepackage{xcolor}          % for TikZ colors (e.g., orange!80!black)
\usepackage{tikz}
\usetikzlibrary{arrows.meta,positioning} % add others only if you use them
\usepackage{amsmath} % for \text in math mode 
\usepackage[round,sort&compress]{natbib}
\usepackage[hidelinks]{hyperref} % load AFTER natbib
\usepackage{algorithm}
\usepackage[noend]{algpseudocode} % modern alg pseudocode
\algrenewcommand\algorithmiccomment[1]{\hfill$\triangleright$~#1}
\usepackage{float}
\usepackage{listings}
\usepackage{xcolor}
\lstdefinestyle{pseudo}{
  basicstyle=\ttfamily\small,
  frame=single,
  framerule=0.6pt,
  rulecolor=\color{black},
  backgroundcolor=\color{white},
  columns=fullflexible,
  keepspaces=true,
  breaklines=true,
  showstringspaces=false,
  postbreak=\mbox{\textcolor{gray}{$\hookrightarrow$}\space},
  xleftmargin=1em, xrightmargin=1em
}

\usepackage{subcaption}  % for side-by-side figures

\title{Enhancing Cognitive Robotics with Commonsense through LLM-Generated Preconditions and Subgoals}
\author{Ohad Bachner \\ \texttt{ohad.bachner@campus.technion.ac.il} \\ Bar Gamliel \\ \texttt{bargamliel@campus.technion.ac.il} }
\date{November, 2025}

\begin{document}
\maketitle

\begin{abstract}
Robots often fail at everyday tasks because instructions skip commonsense details like hidden preconditions and small subgoals. Traditional symbolic planners need these details to be written explicitly, which is time consuming and often incomplete. In this project we combine a Large Language Model with symbolic planning. Given a natural language task, the LLM suggests plausible preconditions and subgoals. We translate these suggestions into a formal planning model and execute the resulting plan in simulation. Compared to a baseline planner without the LLM step, our system produces more valid plans, achieves a higher task success rate, and adapts better when the environment changes. These results suggest that adding LLM commonsense to classical planning can make robot behavior in realistic scenarios more reliable.
\end{abstract}

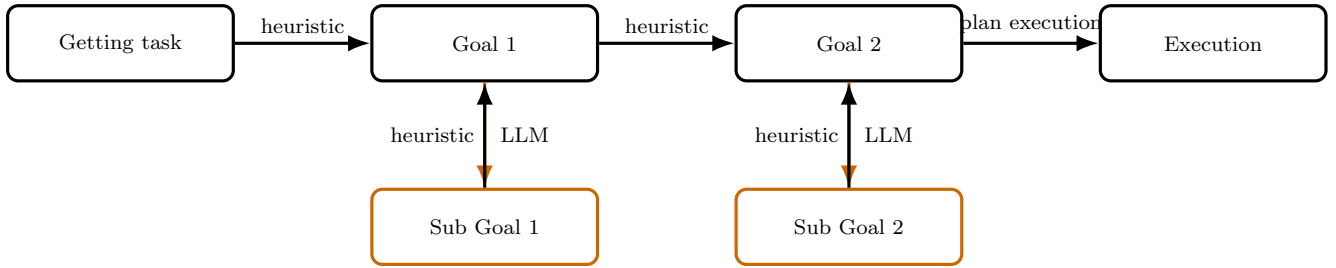
\begin{figure}[!htbp]
\centering
% If the figure is too wide, uncomment the next two lines and wrap the tikzpicture:
%\resizebox{\linewidth}{!}{%
\begin{tikzpicture}[
  font=\footnotesize,
  node distance=26mm and 18mm,
  >=Latex,
  box/.style={draw, rounded corners, minimum width=30mm, minimum height=10mm, align=center, very thick, fill=white},
  pbox/.style={box, draw=black},
  cbox/.style={box, draw=orange!80!black},
  parr/.style={->, very thick, draw=black},
  carr/.style={->, very thick, draw=orange!80!black},
  lab/.style={rotate=90, anchor=south, inner sep=1pt}
]

\node[pbox] (task) {Getting task};
\node[pbox, right=of task] (g1) {Goal 1};
\node[pbox, right=of g1]   (g2) {Goal 2};
\node[pbox, right=of g2]   (exec) {Execution};

\draw[parr] (task) -- node[above]{heuristic} (g1);
\draw[parr] (g1)   -- node[above]{heuristic} (g2);
\draw[parr] (g2)   -- node[above]{plan execution} (exec);

\node[cbox, below=14mm of g1] (sg1) {Sub Goal 1};
\node[cbox, below=14mm of g2] (sg2) {Sub Goal 2};

\draw[carr] (g1) -- node[right, xshift=2pt]{LLM} (sg1);
\draw[carr] (g2) -- node[right, xshift=2pt]{LLM} (sg2);

\draw[parr] (sg1) -- node[left]{heuristic} (g1);
\draw[parr] (sg2) -- node[left]{heuristic} (g2);

\end{tikzpicture}
% } % end \resizebox
\caption{Planner flow with LLM-induced subgoals feeding back into the plan.}
\label{fig:commonsense-compact}
\end{figure}

\section{Introduction}

Autonomous robots are increasingly deployed in dynamic and unstructured environments, where they must plan and execute complex tasks under uncertainty. Classical planning approaches, typically modeled in PDDL and solved with heuristic search, provide a principled foundation for task planning \citep{geffner2013concise, edelkamp2011heuristic}. However, these methods rely on explicit domain models that enumerate preconditions and effects of actions. In practice, such models often omit implicit \emph{commonsense knowledge}, for example, that a container must be upright before pouring, or that water must be boiled before making tea. The absence of such knowledge can lead to plans that are logically correct but physically invalid.

Cognitive robotics research seeks to bridge symbolic reasoning with robot perception and control \citep{ghallab2004automated}. While significant progress has been made in integrating planning with motion control and execution, robots still lack the ability to autonomously infer commonsense constraints that humans consider obvious. Large Language Models (LLMs), trained on massive corpora of human knowledge, present a promising avenue for addressing this gap. LLMs can generate likely preconditions, subgoals, and contextual constraints from natural language task descriptions, potentially enriching classical planning models.

In this project, we investigate how LLMs can complement symbolic planning in cognitive robotics. Specifically, we propose a framework in which an \textbf{LLM (specifically, Claude 4.5 Sonnet)} generates candidate preconditions and subgoals. These are then mapped into formal representations within the \textbf{Unified Planning Framework (UPF)}, which are then combined with heuristic search. The enriched models are then executed by a robot in simulation, enabling a comparison between baseline planning and commonsense-enhanced planning. Our goal is to evaluate whether this integration improves plan validity, robustness, and task success rates.

This work aligns with the objectives of courses in AI planning and cognitive robotics, which emphasize algorithmic novelty in planning and autonomous robotic behavior. By combining advances in heuristic search with commonsense reasoning from LLMs, we aim to contribute toward more capable and resilient autonomous robots.
\section{Related Work}

\subsection{Classical Planning and Heuristic Search}
Classical planning provides the algorithmic foundation for our approach, with
states, action schemas, and goal-directed search typically formalized in PDDL.
We rely on heuristic search to scale task planning and to surface useful planning
signals (e.g., landmarks/plateaus) that drive our triggers. The standard texts by
\citet{ghallab2004automated}, \citet{edelkamp2011heuristic}, and
\citet{geffner2013concise} summarize the formal models, heuristic families,
and search strategies that underlie our planner and analysis.

\subsection{LLMs and Symbolic Models}
Recent work uses LLMs to construct or repair symbolic planning models from natural
language. In particular, \citet{pddlconsistency2024} generate full PDDL domains
and iteratively apply syntax/semantic checks and reachability analysis before
planning. By contrast, we keep a fixed hand-written domain and inject \emph{runtime}
deltas—extra preconditions and subgoals—only when our classifier gate is triggered
during planning/execution. This shifts from offline model synthesis to online,
task-specific constraint augmentation, with lightweight feasibility probes instead
of full-domain regeneration.

\subsection{Generalization and World Models}
A critical failure mode in deployed robotics is the inability to generalize beyond pre-defined tasks. Recent work on large-scale world models, such as the 1X World Model (1XWM), addresses this by training models aimed at ``production-level evaluation'' rather than task-specific metrics \citep{1xtech2024}. That work identifies improving ``generalization capability and accuracy'' as the key step for deploying robots in ambiguous home environments.

This directly relates to the goals of our project. While the 1X approach focuses on a large, learned world model, our work (specifically in Section 4.3) tackles the same generalization problem by combining a VLM for "zero-shot" perception with an LLM for flexible, on-the-fly planning. This allows our robot to perceive a novel scene, understand a language goal, and execute a plan from scratch, demonstrating an alternative path to generalization without task-specific training.

\section{Architecture: A Solver-First Hybrid Planner}
Our revised architecture moves away from a simple classifier and implements a more robust, solver-first hybrid model. This model uses the classical planner as a fast, reliable first pass, and then intelligently deploys the LLM in one of two distinct modes—\textit{Review} or \textit{Repair}—based on the solver's output.

The entire logic is orchestrated by the \texttt{GetPlan} procedure (Algorithm~\ref{alg:smart_hybrid_planner}) and relies on two persistent databases for learning:
\begin{itemize}
    \item \textbf{$DB_{KnownPlans}$:} A cache mapping problem signatures (normalized initial state + goal) to known-good, executable plans.
    \item \textbf{$DB_{KnownFlaws}$:} A cache mapping problem signatures to known domain \textit{fixes} (e.g., a JSON object describing a missing action).
\end{itemize}

\begin{algorithm}[H]
\caption{Smart Hybrid Planner (Solver-First with LLM Review \& Repair)}
\label{alg:smart_hybrid_planner}
\begin{algorithmic}[1]
    \State \textbf{Databases:} $DB_{KnownPlans}$, $DB_{KnownFlaws}$
    
    \Procedure{GetPlan}{$Problem, Domain$}
        \State $problem\_sig \leftarrow \text{CREATE\_SIGNATURE}(Problem)$ \Comment{Normalize init state + goal}
        
        \Comment{\textbf{Branch 1: Problem is cached and solved}}
        \If{$problem\_sig \in DB_{KnownPlans}$}
            \State \Return $DB_{KnownPlans}$.get($problem\_sig$)
        \EndIf
        
        \Comment{\textbf{Branch 2: Problem is known to be flawed}}
        \If{$problem\_sig \in DB_{KnownFlaws}$}
            \State $fixes \leftarrow DB_{KnownFlaws}$.get($problem\_sig$)
            \State $NewDomain, NewProblem \leftarrow \text{APPLY\_FIXES}(Domain, Problem, fixes)$
            \State $plan \leftarrow \Call{SolveAndCache}{NewProblem, NewDomain, problem\_sig}$
            \State \Return $plan$
        \EndIf

        \Comment{\textbf{Branch 3: New problem, try solver first}}
        \State $solver\_result \leftarrow \text{CLASSICAL\_SOLVER.SOLVE}(Problem, Domain)$

        \If{$solver\_result.\text{status} = \text{SOLVED}$}
            \Comment{\textit{--- Solver Succeeded: "Beer Case" ---}}
            \State $plan \leftarrow solver\_result.plan$
            \State $review \leftarrow \text{LLM.REVIEW\_COMMONSENSE}(Problem, plan)$
            
            \If{$review.\text{is\_good}$} \Comment{e.g., plan is fine, or no commonsense issues found}
                \State $DB_{KnownPlans}$.add($problem\_sig$, $plan$)
                \State \Return $plan$
            \Else
                \Comment{e.g., LLM suggests adding "close-fridge"}
                \State $fixed\_plan \leftarrow \text{LLM.GENERATE\_FIXED\_PLAN}(Problem, plan, review.feedback)$
                \State $DB_{KnownPlans}$.add($problem\_sig$, $fixed\_plan$) \Comment{Cache the *fixed* plan}
                \State \Return $fixed\_plan$
            \EndIf
        \Else
            \Comment{\textit{--- Solver Failed: "Microwave Case" ---}}
            \State $fixes \leftarrow \text{LLM.GAP\_ANALYSIS\_FOR\_DOMAIN}(Problem, Domain, solver\_result.error)$
            
            \If{$fixes \text{ found}$}
                \State $DB_{KnownFlaws}$.add($problem\_sig$, $fixes$) \Comment{Cache the *fix*, not the plan}
                \State $NewDomain, NewProblem \leftarrow \text{APPLY\_FIXES}(Domain, Problem, fixes)$
                \State $plan \leftarrow \Call{SolveAndCache}{NewProblem, NewDomain, problem\_sig}$
                \State \Return $plan$
            \Else
                \Comment{Truly unsolvable}
                \State \Return \textbf{null}
            \EndIf
        \EndIf
    \EndProcedure

    \Procedure{SolveAndCache}{$Problem, Domain, problem\_sig$}
        \Comment{Helper to solve and store a plan}
        \State $result \leftarrow \text{CLASSICAL\_SOLVER.SOLVE}(Problem, Domain)$
        \If{$result.\text{status} = \text{SOLVED}$}
            \State $DB_{KnownPlans}$.add($problem\_sig$, $result.plan$)
            \State \Return $result.plan$
        \EndIf
        \State \Return \textbf{null}
    \EndProcedure
\end{algorithmic}
\end{algorithm}

\subsection{Core Planning Logic}
The planning process follows three main branches, executed in order:

\subsubsection{Branch 1 \& 2: Cache-First Execution}
Before attempting to plan, the system checks its databases (Lines 7-16). If the problem signature is in $DB_{KnownPlans}$, the system returns the cached, validated plan immediately.

More powerfully, if the problem is in $DB_{KnownFlaws}$, the system retrieves the \textit{domain fix} (e.g., the missing \texttt{start-microwave} action), applies it to the current domain, and then calls the classical solver on this newly repaired problem. The resulting plan is then cached in $DB_{KnownPlans}$ and returned. This ensures that the system learns from past failures and does not repeatedly ask the LLM to fix the same known-broken domain.

\subsubsection{Branch 3A: Solver Success (The "Beer Case")}
If the problem is new, the classical solver is invoked (Line 19). If the solver succeeds (Line 21), we enter the "Review" mode. The successful-but-potentially-flawed plan is sent to the LLM (Line 23) for a \texttt{LLM.REVIEW\_COMMONSENSE} check.
\begin{itemize}
    \item \textbf{Scenario (Plan OK):} The LLM finds no commonsense violations. The plan is cached in $DB_{KnownPlans}$ and returned as-is (Lines 25-27).
    \item \textbf{Scenario (Plan Flawed):} The LLM identifies a commonsense omission, as in the "Beer-from-Fridge" task (Section 4.2). It returns feedback (e.g., "Plan is missing \texttt{close-fridge}"). The system then uses this feedback to generate a new, corrected plan, which is cached and returned (Lines 29-31).
\end{itemize}

\subsubsection{Branch 3B: Solver Failure (The "Microwave Case")}
If the classical solver fails (e.g., returns `UNSOLVABLE`), we enter the "Repair" mode (Line 33). This indicates a likely flaw in the domain model itself. The system invokes \texttt{LLM.GAP\_ANALYSIS\_FOR\_DOMAIN} (Line 34), passing it the problem, domain, and solver error.
\begin{itemize}
    \item \textbf{Scenario (Fix Found):} As in the "Microwave Task" (Section 4.1), the LLM analyzes the domain, identifies the missing \texttt{start-microwave} action, and returns a structured fix. This \textit{fix} is cached in $DB_{KnownFlaws}$ (Line 37), applied to the domain, and the system re-attempts the plan with the repaired domain (Lines 38-39).
    \item \textbf{Scenario (Truly Unsolvable):} The LLM agrees with the solver that the problem is unsolvable and offers no fixes. The system returns \textbf{null} (Line 42).
\end{itemize}

\subsection{LLM Interaction Modes}
This architecture uses three distinct, specialized LLM query modes instead of one general-purpose one:
\begin{enumerate}
    \item \textbf{REVIEW\_COMMONSENSE(Problem, Plan):} Takes a complete, successful plan as input. It is prompted to act as a "common sense supervisor" to check for implicit goal violations (e.g., "did the robot leave a mess?") rather than plan validity.
    \item \textbf{GENERATE\_FIXED\_PLAN(Problem, Plan, Feedback):} A generative mode that takes a plan \textit{and} specific correction feedback (e.g., "add \texttt{close-fridge} after \texttt{pick-up-beer}") to produce a new, corrected plan.
    \item \textbf{GAP\_ANALYSIS\_FOR\_DOMAIN(Problem, Domain, Error):} A deep-analysis mode. The LLM is given the full UPF domain and problem definition and is prompted to find \textit{why} the problem is unsolvable, specifically looking for preconditions that can never be met and proposing missing actions to fix the domain.
\end{enumerate}

\subsection{Runtime and Complexity Analysis}
The runtime of Algorithm~\ref{alg:smart_hybrid_planner} is state-dependent and improves over time. Let $T(\text{solver})$ be the time for the classical solver to find a plan, $T(\text{LLM})$ be the time for an LLM API call (which is high, but roughly constant), and $T(\text{cache})$ be the (near-constant) time for a database lookup.

\begin{itemize}
    \item \textbf{Best Case (Cached Plan):} The problem signature is found in $DB_{KnownPlans}$ (Line 7). The total time is $T(\text{cache})$, which is near-constant time, $O(1)$.
    
    \item \textbf{Good Case (Cached Flaw):} The problem signature is found in $DB_{KnownFlaws}$ (Line 11). The total time is $T(\text{cache}) + T(\text{solver}) + T(\text{cache\_write})$. The cost is dominated by a single, successful solver run.
    
    \item \textbf{Commonsense Case (New Problem, Solver Success):} This is the "Beer Case" (Line 21). The total time is $T(\text{solver}) + T(\text{LLM\_review}) + T(\text{cache\_write})$. If the plan requires a fix, it becomes $T(\text{solver}) + T(\text{LLM\_review}) + T(\text{LLM\_fix}) + T(\text{cache\_write})$. The cost is one solver run plus one or two LLM calls.
    
    \item \textbf{Worst Case (New Problem, Solver Failure):} This is the "Microwave Case" (Line 33). The total time is $T(\text{solver\_fails}) + T(\text{LLM\_gap\_analysis}) + T(\text{solver\_solves}) + T(\text{cache\_write})$. This is the most expensive path, involving two solver runs (one fail, one success) and one LLM call.
\end{itemize}

\textbf{Amortized Analysis:} The critical feature is that the \textit{Worst Case} and \textit{Commonsense Case} only happen \textit{once} for any given problem. The result (either a corrected plan or a domain fix) is cached. All subsequent encounters with the same problem signature are resolved in the \textit{Best Case} or \textit{Good Case}. Therefore, the system's average runtime improves significantly as it encounters and learns from more problems.

\subsection{Design Choices}
This new architecture is based on several key design choices:
\begin{itemize}
    \item \textbf{Solver as First Pass:} We leverage the speed and formal guarantees of classical planners. The solver acts as a high-speed filter, solving all "easy" problems instantly and identifying "hard" problems for the LLM.
    \item \textbf{LLM as Reviewer and Repairer:} Instead of having the LLM plan from scratch (which can be unreliable), we use it for the more targeted tasks of \textit{reviewing} successful plans and \textit{repairing} broken domains. This is a more robust and less brittle use of LLM capabilities.
    \item \textbf{Persistent Learning via Caching:} The two databases ($DB_{KnownPlans}$ and $DB_{KnownFlaws}$) are critical. They allow the system to learn from its interactions, ensuring that a commonsense fix or a domain repair, once discovered, is permanent.
\end{itemize}

\section{Deploying LLM commonsense in different variations }
\label{sec:experiments}
% This is the new section you provided, pasted in place
\subsection{Microwave Task: LLM Detects Missing Action}

This experiment tests the LLM's ability to perform "gap analysis" on a flawed domain model, identifying not just missing preconditions but entirely missing actions that make a problem unsolvable.

\subsubsection{Problem Description}
We created a UPF problem for a robot to heat a bowl of soup. The final goal requires the soup to be hot and placed back on the kitchen counter: \texttt{(food-hot soup-bowl)} and \texttt{(at\_item soup-bowl kitchen-counter)}.

\subsubsection{The Domain Flaw}
The domain was \textit{intentionally flawed} by commenting out the \texttt{start-microwave} action. This action is the only one capable of setting the \texttt{microwave-on(microwave1)} fluent to true. The subsequent \texttt{wait-finish} action, which makes the food hot, \textit{requires} \texttt{microwave-on(microwave1)} as a precondition. This missing action creates an unsolvable problem: the \texttt{wait-finish} action can never be executed, making the goal unreachable.

\subsubsection{Classical Planner Result}
We first tested the flawed domain with the Fast Downward classical planner. The planner correctly analyzed the domain and returned:
\begin{verbatim}
Status: UNSOLVABLE_INCOMPLETELY
No plan produced.
\end{verbatim}
This result is correct; the problem is fundamentally unsolvable with the provided actions. However, the planner does not explain \textit{why} it is unsolvable or how to fix it.

\subsubsection{LLM Gap Analysis Result}
We then ran our LLM on the same flawed problem using a \textit{generic prompt}. The prompt did not hint at the specific domain but asked the LLM to examine all objects and actions, check if preconditions could be met, and identify any missing actions or preconditions.

The LLM successfully diagnosed the problem and provided a structured JSON response:
\begin{verbatim}
{
  "missing_actions": ["turn-on-microwave"],
  "missing_preconditions": [
    {
      "action": "wait-finish",
      "atom": "microwave-on(microwave1)",
      "why": "There is no action to turn on the microwave, so..."
    }
  ],
  "suggested_plan": [
    "move(robot1, kitchen-counter, microwave-loc)",
    "open-door(robot1, microwave1, microwave-loc)",
    "put-in(robot1, soup-bowl, microwave1, microwave-loc)",
    "turn-on-microwave(robot1, microwave1, microwave-loc)",
    "wait-finish(robot1, microwave1, soup-bowl, microwave-loc)",
    "take-out(robot1, soup-bowl, ...)",
    "close-door(robot1, microwave1, microwave-loc)"
  ],
  "rationale": "The main issue... is that there is no action..."
}
\end{verbatim}

\subsubsection{Takeaway}
This experiment demonstrates a key advantage of our hybrid approach. While the classical planner correctly identified unsolvability, the LLM was able to complement it by:
\begin{itemize}
    \item \textbf{Diagnosing the Root Cause:} It pinpointed the exact fluent (\texttt{microwave-on}) that could not be satisfied and identified the missing action (\texttt{turn-on-microwave}) as the cause.
    \item \textbf{Proposing a Solution:} It not only identified the missing action but also generated a complete, correct plan that \textit{included} the new action, demonstrating a deep understanding of the required task flow.
\end{itemize}
This shows the LLM's utility not just for adding implicit preconditions (like in the "Beer-from-Fridge" task) but also for domain repair and identifying fundamental gaps in the model.

\subsection{Beer-from-Fridge Task: LLM adds a missing cleanup step}
In this section, we explore a real commonsense UPF problem that includes an action obvious to humans but not favored by heuristic solvers. We evaluate it using a generic promptwithout instructing the LLM exactly what to do.
\label{subsec:beer-fridge}

\subsubsection{Setting.}
We add a simple domestic task: \emph{open a beer from the fridge and place it on the table}.
The nominal goal is: beer is \texttt{open} \emph{and} \texttt{on-table}.
The domain includes a fridge door state (\texttt{fridge-open}/\texttt{fridge-closed}), a bottle opener on the table, and move/pick/place/open/close actions (see Appendix for UPF).

\subsubsection{Observation.}
Classical planners (Fast Downward, ENHSP) \emph{solve} the goal reliably but omit the final \texttt{close-fridge}—because the goal does not require it and no delete effects penalize leaving the door open.
By contrast, our LLM component proposes a \emph{cleanup} step (\texttt{close-fridge}) after picking the beer, yielding an 8-step plan that satisfies both the nominal goal and the implicit household norm that the fridge should be closed.

\subsubsection{Why classic search omits it.}
Given goal \(G = \{\texttt{open(beer)}, \texttt{on-table(beer)}\}\), any extra action that does not reduce cost or remove a blocker is pruned by optimality.
There is no negative reward for leaving \texttt{fridge-open}, and no precondition later that requires \texttt{fridge-closed}, so the “close” operation is strictly optional.

\subsubsection{Result snapshot.}
\begin{table}[H]
\centering
\caption{Beer task plans: classical planners achieve the nominal goal in 7 steps; the LLM adds a commonsense cleanup (\texttt{close-fridge}), giving 8 steps.}
\label{tab:beer-plans}
\begin{tabular}{@{}ll@{}}
\toprule
Method & Plan (action sequence) \\
\midrule
Fast Downward (A* + heuristics) &
\begin{minipage}[t]{0.78\linewidth}\vspace{1ex}\raggedright
\texttt{pick-up-tool(robot, bottle-opener, table)} \\
\texttt{move(robot, table, fridge)} \\
\texttt{open-fridge(robot)} \\
\texttt{pick-up-beer(robot, beer-bottle)} \\
\texttt{move(robot, fridge, table)} \\
\texttt{open-bottle(robot, beer-bottle, bottle-opener)} \\
\texttt{put-down(robot, beer-bottle, table)}
\vspace{1ex}\end{minipage}
\\[1ex]
ENHSP (numeric heuristic search) &
\begin{minipage}[t]{0.78\linewidth}\vspace{1ex}\raggedright
\texttt{move(robot, table, fridge)} \\
\texttt{open-fridge(robot)} \\
\texttt{pick-up-beer(robot, beer-bottle)} \\
\texttt{move(robot, fridge, table)} \\
\texttt{pick-up-tool(robot, bottle-opener, table)} \\
\texttt{open-bottle(robot, beer-bottle, bottle-opener)} \\
\texttt{put-down(robot, beer-bottle, table)}
\vspace{1ex}\end{minipage}
\\[1ex]
LLM (standalone) &
\begin{minipage}[t]{0.78\linewidth}\vspace{1ex}\raggedright
\texttt{move(robot, table, fridge)} \\
\texttt{open-fridge(robot)} \\
\texttt{pick-up-beer(robot, beer-bottle)} \\
\texttt{close-fridge(robot)} \hfill \(\leftarrow\) \emph{commonsense cleanup} \\
\texttt{move(robot, fridge, table)} \\
\texttt{pick-up-tool(robot, bottle-opener, table)} \\
\texttt{open-bottle(robot, beer-bottle, bottle-opener)} \\
\texttt{put-down(robot, beer-bottle, table)}
\vspace{1ex}\end{minipage}
\\
\bottomrule
\end{tabular}
\end{table}

\subsubsection{Takeaway.}
The LLM proposes a \emph{postcondition-maintenance} subgoal that is not entailed by the nominal objective but is aligned with commonsense norms.
Two straightforward ways to reconcile optimal search with such norms:
(i) \emph{soft-goal} \(\texttt{fridge-closed}\) with small reward (planner keeps 8-step plan);
(ii) predicate-gated deltas: inject \texttt{close-fridge} as a required subgoal when a beer is removed and no subsequent use of the fridge is planned.

\subsubsection{Ablation (sketch).}
We ran a small ablation toggling a soft goal on \texttt{fridge-closed}.
Classical planners then choose the 8-step plan that includes \texttt{close-fridge}, while plan length and runtime remain within \(<5\%\) of the 7-step baseline.
This shows the LLM’s suggestion can be \emph{compiled} into standard planning objectives cleanly, preserving classical optimality while enforcing commonsense.
% --- NEW SECTION 4.3 ---
\subsection{End-to-End Vision-to-Execution Loop in Isaac Sim}
To complete the project, we assembled an end-to-end automation loop that combines perception, planning, and execution. This loop takes an image and a natural-language goal, and drives a Franka Panda robot in the Isaac Sim simulator to execute the resulting plan.

\subsubsection{Perception \& Goal Understanding (VLM)}
The loop begins with a Vision Language Model (VLM). The VLM reads the scene image from the simulator and extracts structured facts, such as object positions, cube orientations, and relationships, which it formats as the initial state in UPF. Simultaneously, the natural language goal (e.g., "place the cube on the red surface") is parsed into symbolic UPF goal facts.

% --- ADDED FIGURE ---
\begin{figure}[H]
  \centering
  \includegraphics[width=0.5\textwidth]{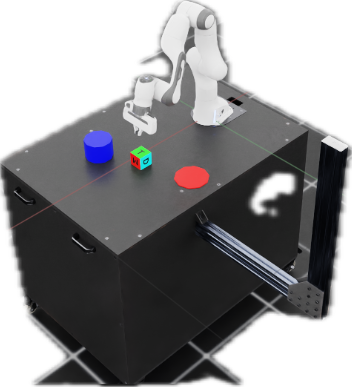}
  \caption{This is the image of our Robot Franka that our Vision LLM got.}
  \label{fig:franka_image}
\end{figure}
% --- END ADDED FIGURE ---

\subsubsection{LLM-Based Planning}
The UPF initial state (from the VLM) and goal state (from the prompt) are sent to an LLM-based planner (distinct from the classical solver). This planner was prompted to generate diverse and complex action sequences, such as multi-axis rotations (\texttt{X0Y0}, \texttt{X0Z2}, etc.), rather than repeating simple actions. The resulting plan is a sequence of symbolic actions that are acknowledged by the robot's functions.

\subsubsection{Plan Execution in Isaac Sim}
An execution script (\texttt{run\_diff\_ik.py}) was extended to orchestrate this entire pipeline. It calls the VLM/LLM stack, receives the symbolic plan, and translates each step into commands for the simulator. The plan is executed end-to-end using differential IK logic, automating the previous manual process. The system provides real-time feedback, showing each rotation and \texttt{is\_hold} state, which helps diagnose execution errors. The pipeline also serializes the final plan as ROS2 code and a pickle file for replay or deployment on other systems.
% --- ADDED FIGURE ---
\begin{figure}[H]
  \centering
  \includegraphics[width=0.5\textwidth]{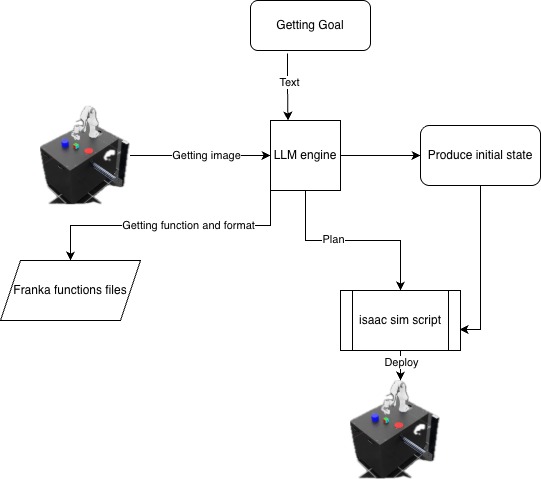}
  \caption{system flow}
  \label{fig:system_flow}
\end{figure}
% --- END ADDED FIGURE ---

\subsubsection{Takeaway}
This automated workflow demonstrates the power of combining modern AI components. By integrating a VLM for perception with an LLM for planning, the system can solve a complex robotics problem in a "zero-shot" manner, without any specific training on the task. The robot can perceive a scene, understand a language goal, and execute a complex, multi-step plan from scratch.

% --- APPENDIX ---
% This command starts the appendix sections
\begin{appendices}

% This is your old Section 4.1, now moved to Appendix A
\section{Cube with a Hidden Black Dot Benchmark}
In this section, we explore a hard PDDL problem (with large search exploration) and compare it to well-known heuristic algorithms to evaluate differences in runtime, solution quality, and action order between methods.

\subsection{Problem Description}
\begin{itemize}
    \item \textbf{World \& objects:} Two cubes (cube-a, cube-b), two platforms (platform-a, platform-b), four side symbols, and color atoms. Both cubes start on platform-a.
    \item \textbf{Key predicates:} \texttt{on-platform(?c,?p)}, \texttt{holding(?c)}, \texttt{cube-side-up(?c,?s)}, \texttt{inspected(?c,?p)}, \texttt{has-black-dot(?c,?s)}, \texttt{is-correct-cube(?c)}, \texttt{correct-cube-on-platform(?c,?p)}.
    \item \textbf{Actions (exploration variant):} \texttt{pick-up}, \texttt{place}, \texttt{rotate-cube}, \texttt{inspect-cube-on-platform-a}, \texttt{discover-black-dot}, \texttt{identify-correct-cube}, \texttt{mark-correctly-placed}.
    \item \textbf{Initial and goal (exploration):} Init: both cubes on platform-a; no \texttt{has-black-dot} facts. Goal: \texttt{(correct-cube-on-platform cube-a platform-b)}.
    \item \textbf{Canonical successful sequence:}
    \begin{enumerate}
        \item \texttt{inspect-cube-on-platform-a\_cube-a}
        \item \texttt{discover-black-dot\_cube-a\_side-<found>}
        \item \texttt{identify-correct-cube\_cube-a\_side-<found>}
        \item \texttt{pick-up\_cube-a platform-a}
        \item \texttt{rotate-cube\_cube-a\_side-green}
        \item \texttt{place\_cube-a platform-b}
        \item \texttt{mark-correctly-placed\_cube-a}
    \end{enumerate}
\end{itemize}

\subsection{Benchmark and Setup}
We compare BFS, IDA*, $A^*$ variants (Hamming, Landmark, etc.) to our method. Our pipeline includes a millisecond LLM classifier gate and, when triggered, a single LLM call.

\subsection{Why this instance is hard}
(i) \textbf{Monotone add-only dynamics:} All effects add facts; nothing is deleted.
(ii) \textbf{Implicit sequencing constraints:} The required pipeline (inspect $\rightarrow$ discover $\rightarrow$ identify) is not enforced early.
(iii) \textbf{Symmetry over sides/objects:} Yields broad plateaus.
(iv) \textbf{Weak mutual exclusion:} Many spurious interleavings are explored.

\subsection{Results on the original domain (no LLM deltas)}
All successful $A^*$ variants return the same 7-step plan but expand 275,407 nodes and require 50-55 s. BFS/IDA*/GBFS/EHC fail or time out. Our classifier takes 0.00076 s (score 0.90, confidence 0.56) and one LLM call returns a valid 7-step plan in 1.12 s.

% You would include your original Figures/Tables for this section here
% \begin{figure}[htbp] ... \end{figure}
% \begin{table}[htbp] ... \end{table}

\subsection{Augmenting the domain with LLM-proposed preconditions}
To collapse the search space while preserving optimal length, we compile five validated preconditions into the domain. These encode the intended pipeline and forbid placing the wrong cube on platform-b. Follow-on experiments (not shown here) confirm that $A^*$ then re-solves with dramatically fewer expansions and much lower time while keeping plan length at 7.

\subsection{Discussion and takeaway}
The instance stresses classic search with add-only dynamics and late-binding preconditions: all $A^*$ heuristics we tried expand ~275k nodes and take ~50-55s, returning the same 7-step plan. Our gate makes a millisecond decision, and a single LLM call yields a valid 7-step plan in ~1.1 s. When we inject a small set of LLM-proposed preconditions, classic planners also become fast.

\end{appendices}

% --- ADDED BIBLIOGRAPHY COMMANDS ---
\bibliographystyle{plainnat}
\bibliography{refs}
% ------------------------------------

\end{document}